\documentclass{sig-alternate}
\newfont{\mycrnotice}{ptmr8t at 7pt}
\newfont{\myconfname}{ptmri8t at 7pt}
\let\crnotice\mycrnotice%
\let\confname\myconfname%

\usepackage{booktabs}
\usepackage[german,american]{babel}
\usepackage[T1]{fontenc}
\usepackage[utf8]{inputenc}
\usepackage{times}
\usepackage{textcomp}
\usepackage{amssymb}
\usepackage{microtype} 
\usepackage{anyfontsize}

\usepackage{url}
\renewcommand{\tt}{\fontencoding{OT1}\fontfamily{cmtt}\selectfont}

\usepackage{graphicx}
\DeclareGraphicsExtensions{.pdf,.ai,.jpg,.png}
\graphicspath{{./}}

\newcommand{\bsfigure}[3][scale=1.0]{%
  \begin{figure}[tb]
    \centering
    \includegraphics[#1]{#2}
    \vspace{-4ex}
    \caption{#3}\label{#2}
    \vspace{-2ex}
  \end{figure}}

\usepackage[usenames,dvipsnames]{color}
\usepackage{booktabs}
\usepackage[numbers,sectionbib]{natbib}
\def\sharedaffiliation{%
\end{tabular}
\begin{tabular}{c}}

\makeatletter
\def\@copyrightspace{\relax}
\makeatother

\newif\ifbscomment
\bscommentfalse
\bscommenttrue

\RequirePackage{soul}
\setstcolor{blue}

\begin{document}

\title{Heuristic Feature Selection for Clickbait Detection}
\subtitle{The Icarfish Clickbait Detector at the Clickbait Challenge~2017}

\def\sharedaffiliation{%
\end{tabular}
\begin{tabular}{c}}

\numberofauthors{1}
\author{
 \alignauthor{Matti Wiegmann\quad Michael V{\"o}lske\quad Benno Stein\quad Matthias Hagen\quad Martin Potthast\vspace*{1ex}}
 \sharedaffiliation
 \affaddr{Bauhaus-Universit{\"a}t Weimar\quad and\quad Leipzig University}\\
 \email{\fontsize{11pt}{12pt}\selectfont <first name>.<last name>@uni-weimar.de\quad and\quad martin.potthast@uni-leipzig.de}\\
}

\maketitle

\begin{abstract}
We study feature selection as a means to optimize the baseline clickbait detector employed at the Clickbait Challenge~2017~\cite{potthast:2017a}. The challenge's task is to score the ``clickbaitiness'' of a given Twitter tweet on a scale from~0 (no clickbait) to~1 (strong clickbait). Unlike most other approaches submitted to the challenge, the baseline approach is based on manual feature engineering and does not compete out of the box with many of the deep learning-based approaches. We show that scaling up feature selection efforts to heuristically identify better-performing feature subsets catapults the performance of the baseline classifier to second rank overall, beating 12~other competing approaches and improving over the baseline performance by~20\%. This demonstrates that traditional classification approaches can still keep up with deep learning on this task.
\end{abstract}

\section{Introduction}

The widespread usage of social media in politics, online news publishing, and public relations gives rise to misuse and abuse in the forms of fake news, hate speech, and clickbait. Especially the high frequency with which new posts are spread on social media causes publishers to compete for user attention: a hunt for clicks. This competition incentivizes the use of clickbait headlines or clickbaiting language to generate page visits. This pertains particularly to the news domain, where publishers seem to balance a certain degree of clickbaitiness against perceived credibility. This balance is a central problem for automated detection: since there is no closed definition that clearly distinguishes clickbait messages from others, the degree of "clickbaitiness" must be measured.

Preceding the Clickbait Challenge~2017, \citet{potthast:2016} suggested an approach for binary clickbait classification using a range of different features at the tweet level and at the document level. This approach has been reimplemented as close as possible to the full feature set of the original publication, but using a regression model instead of a linear classifier in order to serve as a baseline for the challenge.%
\footnote{\url{https://github.com/clickbait-challenge/clickbait17-baseline}}
This baseline achieved the 7th~rank among 13~participants. In their original paper, \citeauthor{potthast:2016} were able to improve the performance of their classifier by~10\% when selecting the 1,000~best features ($\chi^2$ feature selection) from the full feature set. However, adopting the same feature selection for the regression task of the Clickbait Challenge did not yield the same improvement. It appears that $\chi^2$ feature selection does not seem to work as well for scoring clickbaitiness as it does for classification. In this paper, we demonstrate how to improve the performance of the Clickbait Challenge~2017 baseline via a new heuristic approach called leave-many-out feature selection; the Icarfish clickbait detector.%
\footnote{https://github.com/clickbait-challenge/icarfish}

\section{Background and Related Work}

This section briefly reviews related work on clickbait detection, and it gives a detailed background of feature selection.

\subsection{Clickbait Detection}

Our approach derives from the clickbait detection classifier developed by \citet{potthast:2016}, which utilizes 215~different features and feature types, grouped into the following 5~categories: tweet-based features (e.g., character and word n-grams), Downworthy rule sets, General Inquirer word lists, features based on the linked web page, and meta information (e.g., publisher). It was trained on the Webis Clickbait~2016 corpus~\cite{potthast:2016}, a manually annotated 3000~tweet corpus constructed alongside the classifier. Some of the originally proposed features were omitted to render the approach compatible with the challenge's training dataset and the TIRA evaluation platform~\cite{potthast:2014}. For instance, the publisher and retweet information originally used are not present in the challenge's datasets, and the Imagga image tagging service used cannot be reached from within the challenge's TIRA evaluation platform~\cite{potthast:2014}. Furthermore, the Downworthy clickbait rules were omitted, since they are redundant.

The training dataset of the Clickbait Challenge~2017 comprises 19,538~tweets, each annotated by at least 5~people recruited at Amazon's Mechanical Turk~\cite{potthast:2017b}. Another 18,979~tweets have been annotated the same way, but withheld for testing. An overview of all other approaches submitted to the challenge can be found in~\cite{potthast:2017a}, which also includes a review of other related work, so that we can omit a complete review here.

\subsection{Feature Selection}

The performance of machine learning models heavily relies on a good selection of features since the features determine the information that can be used by the learning algorithm. Features can be discriminative (reducing the prediction error), confusing (increasing the prediction error), or redundant (no direct impact). Likewise, feature combinations can have these properties. A good feature selection approach should identify confusing features (or feature combinations) and redundant features to avoid the curse of dimensionality. \citet{Guyon:2003} distinguish three basic strategies for feature selection:
\begin{itemize}
\setlength{\itemsep}{0ex}
\item
{\em Filter methods} estimate the value of a feature by statistical analyses, such as correlation and mutual information.
\item
{\em Embedding methods} integrate filter methods within a machine learning algorithm (e.g., random forest regression).
\item
{\em Wrapper methods} are meta-heuristics that apply a search strategy to find a subset and evaluate it by training a model (e.g., simulated annealing or genetic algorithms with the prediction error as their fitness function).
\end{itemize}

Since the Clickbait Challenge~2017 data are sparse, filter methods and embedding methods are not very effective: statistics like correlation or mutual information cannot provide meaningful information for sparse data as indicated, for example, by the high prediction errors of a lasso regression (considering covariance to select features) or a random forest regression (building decision trees) shown in Figure~\ref{plot-regression-algorithm-over-mean-squared-error}. Heuristic wrapper methods is hence the only alternative left, incurring a comparably high computational complexity.

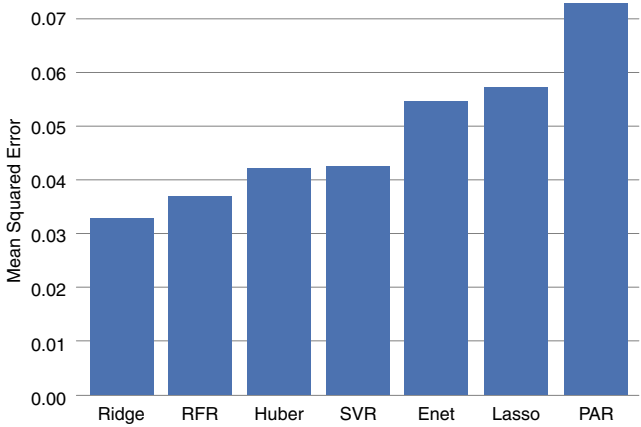
\bsfigure[width=\columnwidth]{plot-regression-algorithm-over-mean-squared-error}{Mean squared errors of out-of-the-box regression algorithms on the Clickbait Challenge 2017 training dataset using the full set of features described in Section~\ref{sec:features}. Shown are ridge, random forest (RFR), Huber, support vector (SVR), elastic net (Enet), lasso, and passive aggressive regression (PAR).}

The fitness function of heuristic feature selection approaches is typically the prediction error; in case of the Clickbait Challenge~2017, the mean squared error (MSE) between the predicted clickbaitiness scores and the ones in the ground-truth is used. In our study, we use the mean squared error as the fitness function to train a ridge regression model,%
\footnote{We used the scikit-learn 0.18.1 library and Python 3.5.2.}
since it is a fast linear model that does not preselect features based on their properties, and since it provides better results using the full feature set compared to most other regression approaches (see Figure~\ref{plot-regression-algorithm-over-mean-squared-error}).

Several local search algorithms and combinatorial optimization algorithms can find good approximations of the ``optimal'' feature subset (e.g., simulated annealing, genetic algorithms, and ant colony optimization). The major downside is that they usually identify a good subset but give no information about how good the features are and how they interact. Determining an individual feature's quality and taking into account feature relationships will yield more insights about the task at hand and might even help to engineer better features. We therefore do not employ simulated annealing, genetic algorithms, or ant colony optimization in our study.

\enlargethispage{\baselineskip}
As noted earlier, feature selection can significantly increase the performance of predictive models by generalizing and reducing computational load for further experiments and analysis. Established strategies usually use statistics such as $\chi^2$ or correlations between a feature and the target vector. When working on short texts, such as social media posts or comments, all features based on occurrence frequencies (like n-grams or word lists) become sparse. Sparse vectors on their own do not encode a lot of information and therefore the statistical metrics become indifferent. This can be seen well when considering the performance of learning algorithms that fit using these metrics (like lasso or elastic net). This also implies that a feature selection strategy is needed which considers combinations (or interactions) of features. 

Considering that statistical metrics are not helpful, applying a search strategy is the logical next step. These strategies iteratively add and/or remove features using prediction error changes for guidance. Established strategies include the following:
\begin{itemize}
\setlength{\itemsep}{0ex}
\item
{\em Exhaustive Search}, which is exponential in the number of features.
\item
{\em Forward Selection}, which does not find interactions.
\item
{\em Compound Selection}, which adds $k$ features and removes $r$, $k > r$. This strategy is computationally feasible if $k - r$ is smaller than the the total number of features, but does not reliably find feature interactions: If the interactions involve many features and if multiple interactions are considered (which is very likely for sparse text data), $k$ must be chosen large (to capture interactions and change the prediction error in a meaningful way) and $r$ has to become dynamic (large if many redundant features are added, small if a useful interaction was added), rendering compound selection approximately equivalent to the following strategy.
\item
{\em Backward Selection} starts with a full feature set and iteratively removes features that change the prediction error below a certain threshold. This strategy is linear in the number of features but does not find multiple interactions.
\end{itemize}

None of these strategies is ``perfect,'' but backward selection is a good starting point once modified to account for the aforementioned characteristics of the data. If one feature contributes equally to confusing as well as discriminatory interactions, removing it would not change the prediction error. To reliably resolve such interactions, it is necessary to remove features in a way most confusing interactions are resolved and most discriminatory ones are still intact. This requires a way to judge each feature's influence on the whole set. 

The most promising strategy to determine the value of a feature is the leave-one-feature-out error minimization strategy~\cite{weston:2000}. This strategy compares the prediction error of two models. The first one trains and predicts on all features. The second one trains and predicts on all features except the feature in question. The difference between both models (prediction error without the feature minus prediction error with the feature) hints at the feature's contribution in the prediction. Here, a positive value indicates that, if a feature is removed, the prediction error rises and vice versa.

To infer useful information about the value of a feature from its leave-one-out error, it is necessary to first remove redundancy and multiple interactions. The rationale is that smaller feature subsets tend to contain less redundancies and interactions and that the leave-one-out error over a subset of features becomes clearer. Considering the inherent ambiguity of language, it is likely unnecessary to use a strategy to construct a subset. It is also likely that, if the subsets are too small, too much information about the interactions is removed. Since we do not know which subsets are useful for measuring the leave-one-out error of a feature, parallel backward selection is used to determine the subsets, resulting in our proposed leave-many-out feature selection strategy.

\section{Icarfish Clickbait Detector}

The most salient property of the Icarfish clickbait detector is its feature selection approach. Based on the feature set of the challenge's baseline detector, we develop a leave-many-out feature selection heuristic, which, when applied at scale, yields significant improvements over the baseline's performance. In what follows, we overview the feature set and describe our feature selection heuristic and its implementation.

\subsection{Features} \label{sec:features}

The following features are extracted from a Tweet's text only:%
\footnote{If necessary for a given feature, tweets are preprocessed using the WordNetLemmatizer and the TweetTokenizer of NLTK~3.2.4.}
\begin{enumerate}
\setlength{\itemsep}{0ex}
\item
{\em Tweet character n-grams}. All character 1-, 2-, and 3-grams that occur more than twice, resulting in a total of 12,471~distinct character n-gram features weighted by tf-idf. N-grams occurring only once or twice are prone to overfitting.
\item
{\em Tweet word n-grams}. All word 1-, 2-, and 3-grams that occur more than twice, resulting in a total of 24,861~distinct word n-gram features weighted by tf-idf.
\item
{\em Engineered features}. In total, twelve engineered features are implemented. Three features encode character counts: average word length, length of the longest word, and total character length. Five features encode character occurrences: the occurrence frequency of '@'~(mentions), '\#'~(hashtags), and '.'~(dots), whether a tweet starts with a number, and the occurrence frequency of abbreviations following the Oxford abbreviations list.%
\footnote{http://public.oed.com/how-to-use-the-oed/abbreviations/}
Two features encode meta-data, namely whether the tweet has media attachments and the part of day (as quarters of a day, 1 to 4) the tweet was issued. The last two features encode the tweet's sentiment polarity as per the VADER sentiment detector implementation in NLTK~\cite{hutto:2014}, and its Flesh-Kincaid readability score.
\item
{\em Word lists}. Given the 181~General Inquirer word lists,%
\footnote{http://wjh.harvard.edu/\textasciitilde inquirer/}
the Terrier stop word list,%
\footnote{\small https://github.com/terrier-org/terrier-core/blob/4.2/share/stopword-list.txt}
the Dale-Chall easy words list~\cite{chall:1995}, and the Downworthy common clickbait phrases,%
\footnote{\small https://github.com/snipe/downworthy/blob/master/Source/dictionaries/original.js}
each word list is used as a single feature, indicating how often any word form the list occurs in a tweet.
\end{enumerate}
Features from the web pages linked in a tweet are omitted for practical reasons: a clickbait scorer which does not have to download the linked web page of a teaser message is more scalable than one which does. Altogether, we obtain 37,528~distinct features.

\subsection{Regression Model}

As a regression model, we employ ridge regression.%
\footnote{Implementation from scikit-learn~0.18.1 in Python~3.5.2.}
It is a fast linear model that does not preselect features based on statistical properties, which is an important prerequisite for our feature selection experiments. Moreover, when trained on a random 7:3~training-validation split of the training data, using the entire set of features, it achieves a reasonable performance of 0.0328~mean squared error (see Figure~\ref{plot-regression-algorithm-over-mean-squared-error}). In what follows, we employ ridge regression on every feature subset analyzed, creating a new random 7:3~training-validation split each time.

\subsection{Leave-Many-Out Feature Selection}

Leave-many-out feature selection repeatedly applies the backward selection search, averaging the results of leave-one-out errors of each feature removed. Starting with the full feature set of size~$n$, features are randomly removed one at a time until a minimum subset size~$m$ is reached. For each removed feature, the leave-one-out error is recorded. This procedure is repeated $r$~times, resulting in a series of leave-one-out errors for each feature. The average over all leave-one-out errors recorded for a given feature can be considered a score of its overall usefulness. This leave-many-out score captures how frequently and by how much the removal of a feature improves or reduces the prediction error of our regression model, or if its removal has no impact. The leave-many-out score also reflects if a feature is more important for some interactions, or if it is redundant. A negative leave-many-out score indicates that a feature affects prediction errors negatively, a zero score indicates a features is redundant, and a positive leave-many-out score indicates that a feature is useful, where higher scores indicate higher usefulness. Finally, ranking features by their leave-many-out scores and selecting the top~$k$ ones results in a highly discriminatory subset of features.

The quality of the leave-many-out score of a feature, and the runtime of the leave-many-out algorithm depends on the number of removals per run $n - m$, and the number of runs~$r$. If the minimum subset size~$m$ is chosen too small, the last features removed will likely break fewer interactions, yielding scores closer to~0. If~$m$ is chosen too large, the reverse effect may occur. Increasing the number of runs~$r$ will calculate more leave-one-out errors per feature and thus provide better leave-many-out scores, but also increase runtime. Pilot experiments on sparse text data showed that choosing~$m$ and~$r$ so that $r(n - m)/n = 25$ provides good results. Choosing the number~$k$ of top-scoring features to be selected varies depending on the application. Removing features with a negative leave-many-out score always improves the prediction, whereas, for the top-scoring features, local optimization can be applied.

\subsection{Hadoop-based Implementation}

We implement leave-many-out feature selection using the Hadoop framework,%
\footnote{Hadoop streaming~2.7.2}
consisting of preprocessing, mappers, and reducers. The mappers configure the reducers by passing feature subset changes to them, and the reducers conduct the steps training and evaluation in parallel on given subsets, recording the mean squared error changes.

Preprocessing includes parsing the provided dataset, extracting the aforementioned features into a feature matrix, as well as filling a vector with the ground-truth clickbaitiness scores. This data is stored as two NPZ-files (Numpys data storage format). Afterwards, the MapReduce input file is generated. This file dictates how many models are analyzed in parallel (1000~in our case): each line consists of a unique~ID and the size of the full feature set (37,528). Both NPZ-files and the input file are then fed to a Hadoop streaming job.

The mapper reads its input file split one line at a time, generating a bit set of the given length with all bits set to~1, where the $i$-th bit corresponds to the $i$-th feature, its truth value indicating whether the feature is to be included in a to-be-analyzed feature subset. It then repeatedly flips one of the 1-bits to 0, passing the bit set resulting from each bit flip to a reducer, keyed by the line's ID (1000~repetitions in our case). Altogether, for each of the 1000~input lines, our mapper emits 1000~bit sets to the reducers, ordered by ID and consecutive feature removals, resulting in an average 26~removals of each feature overall.

We adjusted the Hadoop job to spawn one reducer per line~ID. Since reducers are typically spawned only once per line~ID, they need to initialize the regression model and the supplied NPZ-files only once before handling the series of bit sets for a given~ID. To ensure generalizability and to foreclose overfitting, each reducer randomly splits the training data into 7:3~training-validation sets. It then proceeds to handle the bit sets one at a time, computing the mean squared error of the regression model after removing the respective features indicated by a bit set, and emitting the recorded prediction error along the bit set as a reference which feature subset was used to compute it, keyed by the input line's~ID.

\section{Evaluation}

We conducted a large-scale experiment to select the ``best'' feature set based on the training dataset of the Clickbait Challenge~2017. The resulting model was then evaluated once on the challenge's test dataset. Furthermore, we analyzed the individual features selected to gain insights about the impact of the best and worst features.

\subsection{Selecting the ``Best'' Feature Set}

In total, we processed one~million remove-train-predict iterations for our feature assessment (i.e., a feature was removed from an average of 26~feature subsets). After the outputs of the iterations were accumulated, the leave-many-out scores per feature was computed to rank the features. From this ranking, we then choose feature subsets of increasing size to examine the effect of removing the lowest ranking features and training a model on the remaining ones. We constructed 49~subsets including the top-ranked 100\%,~\dots,~2\% of the features in steps of~2\%, and three other subsets containing only the top-ranked 1.5\%, 1.0\%, and 0.5\% of the features. For every subset, we trained a ridge regression model on 2/3~of the training data and validated it on the remaining 1/3~of the training data. In this case the split was kept constant for all the 52~feature subsets. Figure~\ref{plot-feature-set-size-over-mean-squared-error} shows the mean squared errors measured on the 1/3~of the training data used for validation. The best-performing model in this experiment series achieves a mean squared error of~0.0297 with 12,008~features (32.0\%~of the original features). This corresponds to about 10\%~performance improvement over the full feature set (see Table~\ref{table-evaluation-results}, left). The smallest subset size that does not perform worse than the full feature set uses 375~features~(1.0\%).

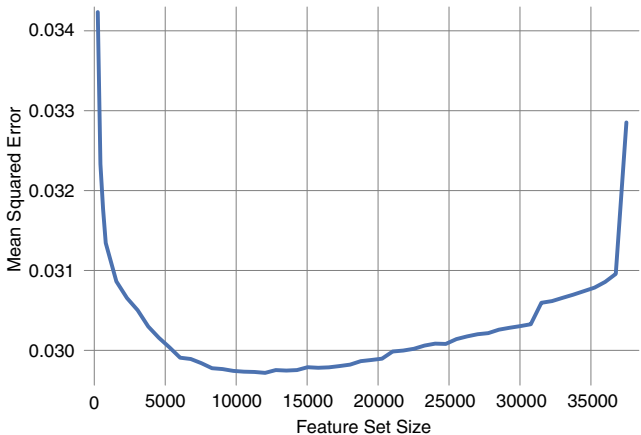
\bsfigure[width=\columnwidth]{plot-feature-set-size-over-mean-squared-error}{Mean squared error of the ridge regression models for 52~feature subsets on a constant 2:1 training-validation-split of the Clickbait Challenge~2017 training dataset.}

We chose the performance-wise best among the validated models as our final approach, and submitted it for testing to the challenge's evaluation platform, achieving a prediction error of~0.0351 on the Clickbait Challenge~2017 test data (participant Icarfish in Table~\ref{table-evaluation-results}, right). This means that Icarfish, based only on its heuristic feature selection, achieves a performance gain of about 20\%~over the baseline and would have been placed second in the challenge.%
\footnote{Icarfish did not officially compete in the challenge.}

\begin{table}[tb]%
\vspace{-2ex}%
\small%
\centering%
\renewcommand{\tabcolsep}{5pt}%
\renewcommand{\arraystretch}{1}%
\caption{Mean squared error (MSE) of selected models on a 2:1~training-validation-split of the Clickbait Challenge~2017 training dataset (left), and means squared error of selected submitted approaches on the challenge's test dataset (right).}%
\label{table-evaluation-results}%
\vspace{1ex}%
\begin{tabular}{@{}rrc@{}}
\toprule
\multicolumn{2}{@{}c}{\bfseries Feature Set Size} & \bfseries MSE\\
\cmidrule(lr@{\tabcolsep}){1-2}
absolute & relative \\
\midrule
37,528 & 100.0\% & 0.0328 \\
12,008 &  32.0\% & 0.0297 \\
   375 &   1.0\% & 0.0323 \\
   187 &   0.5\% & 0.0342 \\
\bottomrule
\end{tabular}%
\hfill%
\renewcommand{\arraystretch}{1.07}%
\begin{tabular}{@{}lcc@{}}
\toprule
\bfseries Participant & \bfseries Rank & \bfseries MSE\\
\midrule
Zingel                          & 1  & 0.0332 \\
\bfseries Icarfish (this paper) & -- & \bfseries 0.0351 \\
Emperor                         & 2  & 0.0359 \\
Carpetshark                     & 3  & 0.0362 \\
Clickbait17-baseline            & 7  & 0.0435 \\
\bottomrule
\end{tabular}%
\vspace{-2ex}%
\end{table}

\subsection{Feature Analysis}

To gain insights into the 52~feature subsets of our experiment, we analyzed them based on the validation data obtained from the challenge's training data. Figure~\ref{plot-feature-set-size-over-mean-squared-error} shows that removing the first~2\% of the most ``confusing'' features provides by far the highest performance gain among any neighboring pair of feature subsets. These ``worst'' features could be features that are confusing in themselves or features whose removal resolves confusing interactions. It further appears that about 68\%~of the features have an at least slightly confusing impact, probably due to feature interactions that are resolved by removing one of them. And the reverse also seems to hold: There are some, but far fewer features with beneficial interactions among them, slightly increasing the performance when included.

Figure~\ref{plot-feature-set-over-leave-many-out-score} and Table~\ref{table-top-scoring-features} show the features with high impact (absolute value of performance gain or loss larger than $10^{-5}$) split by feature category. All feature types include good and bad features with differences in strength and frequency. Most of the 12~engineered features have a high impact about one order of magnitude larger than that of any other feature. However, two of the engineered features harm prediction performance (the average word length and whether a tweet starts with a number). One reason for the relatively high impact of the engineered features is that they are not as sparse as the n-grams or word lists. Interestingly, word n-grams are the only feature category without high-impact confusing features. An interesting side note is that the question mark seems to be discriminatory as a word token, but confusing as a character token, while the exclamation mark is discriminatory in both categories.

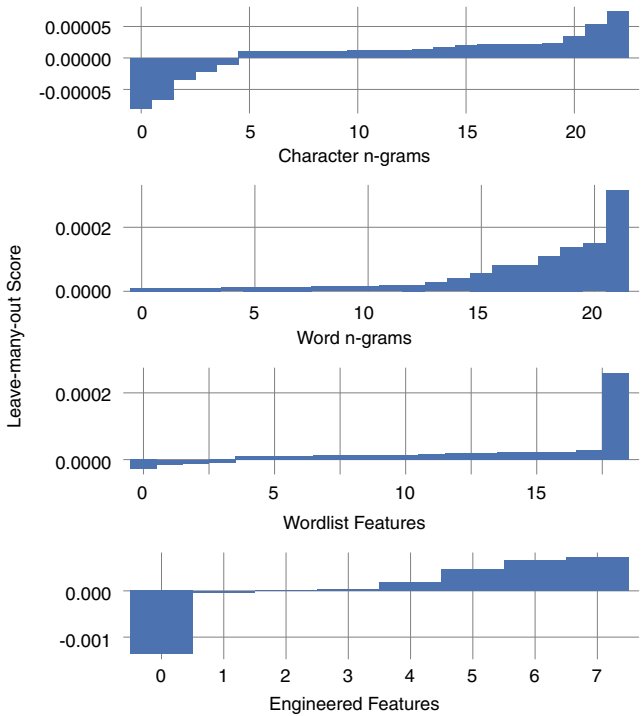
\bsfigure[width=\columnwidth]{plot-feature-set-over-leave-many-out-score}{Leave-many-out scores grouped by feature category and sorted by performance impact. The height of the bar corresponds to the average leave-many-out score of the subsets including the feature. A high positive bar indicates discriminative features. Features with scores below~0.0001 are omitted.}

From Figure~\ref{plot-feature-set-over-leave-many-out-score} and Table~\ref{table-top-scoring-features}, it seems as if the engineered features from the baseline approach are among the most valuable ones for scoring clickbaitiness. Maybe more compact models could thus be constructed if the information of individual features with high impact could be combined into engineered features with higher impact. An example for a new engineered feature could be something like the ``number of words ending with -ly or -ing'' (counting adverbs, gerunds, etc.) since these suffixes seem to be rather discriminating character n-gram features. However, the hypothesis of creating good engineered features from individual other features is only partially supported by a qualitative analysis of the high-impact features. While the engineered ``number of stopwords'' feature is very discriminatory and most of the high-impact word 1-grams are actually stop words, the engineered ``number of mentions'' feature (counting the number of @'s in a tweet) is not in the table of the high-impact engineered features, even though 3~of the 18~most discriminatory character tokens involve the '@'~character.

\begin{table}[tb]%
\vspace{-2ex}%
\fontsize{7.5pt}{8.5pt}\selectfont%
\centering%
\renewcommand{\tabcolsep}{1pt}%
\caption{Features by category with a leave-many-out score $> 10^{-5}$. All scores are multiplied by $10^{-4}$. Feature categories shown are character n-grams (CNG), word n-grams (WNG), word list features, and engineered features.}%
\label{table-top-scoring-features}%
\vspace{1ex}%
\renewcommand{\arraystretch}{0.86}%
\begin{tabular}{@{}lrlr@{}}
\toprule
\bfseries CNG & \bfseries Score \\
\midrule
\tt thi                 & -0.80 \\
\tt his                 & -0.66 \\
\tt ?                   & -0.34 \\
\tt hi                  & -0.21 \\
\tt th                  & -0.11 \\
\tt hes                 &  0.10 \\
\tt ing                 &  0.10 \\
\tt wh                  &  0.11 \\
\tt !                   &  0.11 \\
\tt how                 &  0.11 \\
\tt ly                  &  0.12 \\
\tt ly\textvisiblespace &  0.12 \\
\tt s                   &  0.12 \\
\tt j                   &  0.14 \\
\tt n                   &  0.16 \\
\tt .@                  &  0.20 \\
\tt ..                  &  0.21 \\
\tt s\textvisiblespace  &  0.21 \\
\tt a                   &  0.26 \\
\tt \textvisiblespace @ &  0.24 \\
\tt .                   &  0.34 \\
\tt \textvisiblespace   &  0.53 \\
\tt @                   &  0.73 \\
\bottomrule
\end{tabular}%
\hfill%
\renewcommand{\arraystretch}{0.9}%
\begin{tabular}{@{}lrlr@{}}
\toprule
\bfseries WNG & \bfseries Score \\
\midrule
\tt how to & 0.11 \\
\tt as     & 0.11 \\
\tt sex    & 0.11 \\
\tt dies   & 0.11 \\
\tt 5      & 0.11 \\
\tt that   & 0.13 \\
\tt an     & 0.13 \\
\tt to     & 0.14 \\
\tt \dots  & 0.15 \\
\tt in     & 0.15 \\
\tt here   & 0.15 \\
\tt things & 0.18 \\
\tt heres  & 0.20 \\
\tt is     & 0.28 \\
\tt at     & 0.40 \\
\tt a      & 0.57 \\
\tt !      & 0.82 \\
\tt how    & 0.83 \\
\tt these  & 1.10 \\
\tt ?      & 1.38 \\
\tt .      & 1.50 \\
\tt this   & 3.16 \\
\bottomrule
\end{tabular}%
\hfill%
\renewcommand{\arraystretch}{1.035}%
\begin{tabular}{@{}lr@{}}
\toprule
\bfseries Word lists & \bfseries Score \\
\midrule
RcLoss         & -0.29 \\
Human coll.    & -0.15 \\
PowCon         & -0.12 \\
EnlTot         & -0.11 \\
Exprsv         &  0.11 \\
You            &  0.12 \\
Virtue         &  0.12 \\
PowLoss        &  0.12 \\
Quan           &  0.14 \\
Dist           &  0.15 \\
Vice           &  0.15 \\
PowDoct        &  0.15 \\
Polit          &  0.19 \\
Finish         &  0.20 \\
Academic       &  0.22 \\
Eval           &  0.22 \\
PowEnds        &  0.23 \\
Stop Words     &  0.29 \\
Easy Words     &  2.60 \\
\bottomrule
\end{tabular}%
\hfill%
\renewcommand{\arraystretch}{1.0}%
\begin{tabular}{@{}lrlr@{}}
\toprule
\bfseries Engineered & \bfseries Score \\
\midrule
Mean word length    & -1.37  \\
Starts with number  & -0.30  \\
Abbreviation count  &  0.11  \\
Sentiment polarity  &  0.25  \\
Max. word length    &  1.90  \\
Part of day         &  4.74  \\
Character sum       &  6.70  \\
Number of dots      &  7.18  \\
\bottomrule
\vspace*{27ex}
\end{tabular}%
\vspace{-2ex}%
\end{table}

\section{Conclusion}

Our study has demonstrated that an iterated leave-many-out strategy can help to identify good feature subsets from the ones employed in the Clickbait Challenge~2017 baseline. The challenge poses clickbaitiness scoring as a regression task on social media posts and thus typically comes with rather sparse features---a particularly challenging scenario for feature selection. The best feature subsets that our approach identified achieve a performance gain of~20\% over the baseline with respect to the prediction error. At the same time, our technique is able to hint at possible feature interactions and further feature engineering possibilities.

Concerning the actual subset selection, open questions for future research still are how many different random feature subsets are needed to draw valid conclusions about an individual feature's potential, and how to best cover potentially interacting feature groups in the selection of the removed subsets. Furthermore, our initial observations on different features' strengths may hint at some possible engineered features that combine the individual features' potential---integrating more of such engineered features in a regression model could also be an interesting future direction. Finally, ensemble techniques for feature-based learning or a combination with deep learning are further opportunities.

\begin{raggedright}
\bibliography{clickbait17-notebook-lit}
\end{raggedright}
\end{document}